# Can Large Language Models Capture Public Opinion about Global Warming? An Empirical Assessment of Algorithmic Fidelity and Bias

Sanguk Lee[1],
Tai-Quan Peng[2*],
Matthew H. Goldberg[1],
Seth A. Rosenthal[1],
John E. Kotcher[3],
Edward W. Maibach[3],
Anthony Leiserowitz[1]

1. Yale Program on Climate Change Communication, Yale University, New Haven, CT, USA
2. Department of Communication, Michigan State University, East Lansing, MI, USA
3. Center for Climate Change Communication, George Mason University, Fairfax, VA, USA
[*]Corresponding author: winsonpeng@gmail.com (Dr. Tai-Quan Peng)

## Abstract

Large language models (LLMs) have demonstrated their potential in social science research by emulating human perceptions and behaviors, a concept referred to as algorithmic fidelity. This study assesses the algorithmic fidelity and bias of LLMs by utilizing two nationally representative climate change surveys. The LLMs were conditioned on demographics and/or psychological covariates to simulate survey responses. The findings indicate that LLMs can effectively capture presidential voting behaviors but encounter challenges in accurately representing global warming perspectives when relevant covariates are not included. GPT-4 exhibits improved performance when conditioned on both demographics and covariates. However, disparities emerge in LLM estimations of the views of certain groups, with LLMs tending to underestimate worry about global warming among Black Americans. While highlighting the potential of LLMs to aid social science research, these results underscore the importance of meticulous conditioning, model selection, survey question format, and bias assessment when employing LLMs for survey simulation. Further investigation into prompt engineering and algorithm auditing is essential to harness the power of LLMs while addressing their inherent limitations.

# 1. Introduction

It is very important to measure public opinion about global warming, as these opinions can have considerable influence over policy-making decisions (Bromley-Trujillo & Poe, 2020) and shape public behavior (Doherty & Webler, 2016). A primary method employed by scholars and policymakers for measuring and assessing these opinions is through representative surveys (Berinsky, 2017). However, the extensive time and financial resources required for these surveys can hinder the timely tracking of evolving public opinions about global warming. Resource constraints can also lead to an unintended bias towards majority opinions, potentially neglecting the perspectives of minority groups due to their typically smaller sample sizes in national representative surveys. Nonetheless, understanding diverse public opinion regarding global warming is also vital for climate justice. This understanding can promote equitable decision-making, elevate the concerns of vulnerable communities, help align climate policies with democratic principles, build public support, and address disparities in climate change awareness and priorities. Furthermore, understanding the diversity of public opinion can help support a just transition and mobilize support for climate justice initiatives.

Large Language Models (LLMs) like ChatGPT have the potential to complement traditional survey methods by simulating survey responses with fewer resources and augmenting data from underrepresented sub-populations. Moreover, if LLMs can effectively predict individual opinions on global warming, they could substantially aid researchers in refining research methodologies, for example forecasting outcomes before initiating primary studies. As an initial assessment of the potential of using LLMs for global warming survey research, this study investigates the extent to which LLMs accurately emulate and reflect multiple dimensions of public opinion about global warming.

LLMs have demonstrated their significant potential for contributing to social science research. A recent development lies in their capacity to replicate the perceptions, viewpoints, and behavior of the general population or specific subgroups, termed *algorithmic fidelity* (Argyle et al., 2023). Algorithmic fidelity refers to the extent to which LLMs' intricate web of connections among ideas, attitudes, and sociocultural contexts accurately reflects those found in various human subgroups (Argyle et al., 2023). This emerging approach offers a promising avenue to leverage LLMs in social science, including surveys and experiments. By training on an extensive corpus of human-generated data that includes human perceptions and behaviors, LLMs may possess the capability to simulate diverse facets of public opinions.

Recent studies have yielded promising results. For instance, Argyle et al. (2023) found strikingly high correlations in voting behaviors during presidential elections between human samples and *silicon samples* derived from LLMs. Silicon samples refers to those synthesized by LLMs conditioned to thousands of sociodemographic backstories sourced from real human participants in surveys (Argyle et al., 2023). Similarly, Hwang et al. (2023) found that LLMs were able to accurately reflect public opinions on diverse political issues, including gun control, gender perspectives, economic inequality, trust in science, and so forth. However, the majority of these studies have primarily focused on the political domain, particularly presidential elections and support for political issues.

It remains uncertain whether LLMs can accurately represent public beliefs and emotions about other crucial and prominent social topics, such as global warming. Although perceptions of global warming are driven by subjective and experiential factors, the issue of climate change is grounded in science, which is different from political opinions. Given this distinction, LLMs might exhibit different performance when predicting public perspectives on global warming.

LLMs are engineered to prioritize correctness through extensive training and alignment processes. Empirical evidence suggests that silicon samples generated by LLMs provide hyper accurate responses that are far from human responses when asked about scientific facts such as the melting temperature of aluminum (Aher et al., 2023). The inclination towards correctness could potentially hinder the LLMs' capability to capture the heterogeneous and sometimes inaccurate human perspectives on global warming.

Algorithmic fidelity in LLMs can be affected by several factors, including conditioning inputs and the choice of LLM models. To accurately reflect public opinion on global warming using LLMs, it is essential to provide these models with detailed inputs, such as demographics and covariates. By conditioning LLMs with data reflecting demographic traits and opinions from surveys, they can better portray different individual's opinions regarding particular questions. Prior research has demonstrated that LLMs enhance their predictive accuracy for a wide array of political issues when informed by an individual's past opinions, as opposed to merely leveraging demographics and ideology (Hwang et al., 2023). Furthermore, the specific model version might influence their algorithmic fidelity given the pronounced differences in their capabilities for various intellectual tasks (OpenAI, 2023). Empirical studies also find that algorithmic fidelity varies among LLM models (Aher et al., 2023; Santurkar et al., 2023). Building on these studies, we assess algorithmic fidelity under different conditions and models. Specifically, we compare LLMs conditioned solely on demographics with LLMs conditioned on both demographics and issue-related covariates. For the sake of simplicity, we categorize political ideology and party affiliation as part of demographics. Furthermore, we examine the algorithmic fidelity of distinct LLM versions: GPT-3.5 and GPT-4.

Public perceptions of global warming are complicated, including beliefs about global warming, understanding its causes, emotional responses like worry, policy support, and behavior. To gain deeper insights into how well LLMs accurately represent diverse psychological aspects related to global warming, we assess the evaluation metrics (e.g., accuracy, F1) and distribution of LLM predictions in comparison to survey responses that gauge these various dimensions. Strong performance in these metrics indicates robust algorithmic fidelity, leading to closely matching distributions between LLM-generated samples and actual survey results. The study draws on nationally representative climate change survey data collected in 2017 and 2021 as the benchmark for evaluating the algorithmic fidelity of LLMs.

## 2. Methods

### 2.1. Survey Sampling

Two nationally representative survey datasets were collected in October 2017 (N = 1304) and September 2021 (N = 1006). In these surveys, participants were asked to answer multiple questions related to global warming. These surveys were conducted under an exemption granted by the Institutional Review Board (IRB) of [Redacted] (IRB Protocol ID: [Redacted]). For each survey, the researchers obtained a distinct sample from the Ipsos KnowledgePanel, comprising U.S. adults aged 18 and over. This panel, which mirrors the U.S. population, was assembled using probability sampling methods. Panel members were recruited using various techniques, such as random digit dialing and address-based sampling, covering nearly all residential phone numbers and addresses in the U.S. Participants completed the survey forms online. Those without internet access were provided with computers and internet connectivity.

### 2.2. Silicon Sample Data Collection

To generate silicon sample datasets, we used two versions of GPT (GPT-3.5 vs. GPT-4) and two sets of conditional inputs (demographics only vs. demographics and issue-related covariates). Specifically, silicon samples were generated using GPT-3.5-turbo-16k and GPT-4 through the OpenAI API, setting the temperature at 0.70 based on a prior study (Argyle et al., 2023). For models conditioned solely on demographics, we fed demographic information, such as race/ethnicity, gender, age, political ideology, political party affiliation, education, and residential state, into the models via prompts. Meanwhile, for the models conditioned on both demographics and covariates, additional covariates such as issue involvement in global warming, interpersonal discussions about the topic, and awareness of the scientific consensus, were included along with demographics. These covariates were selected because they appear commonly in both waves of the survey and served as covariates in previous studies (Goldberg et al., 2019; Hornsey et al., 2016; Reser & Bradley, 2020; Van Der Linden et al., 2015).

We utilized an interview format adapted from Argyle et al. (2023) for our prompts (prompt examples are available in the supplemental document). At the system level, GPTs were instructed to act as an interviewee, guided by the directive: "You are an interviewee. Based on your previous answers, respond to the last question." Subsequently, the simulated interview began. To establish a clear timeline for GPTs, the first prompt was phrased as: "Interviewer: What is the current year and month of this interview? Me: October 2017." For the 2021 survey, "October 2017" was replaced with "September 2021." After setting the timeline, "Me" responses leading up to the final question were provided using actual survey data. For instance, regarding race/ethnicity, the prompt was framed as: "Interviewer: I am going to read you a list of five race categories. What race do you consider yourself to be? 'White, Non-Hispanic', 'Black, Non-

Hispanic', '2+ Races, Non-Hispanic', 'Hispanic', or 'Other, Non-Hispanic.' Me: {race from survey response}."

The final question is the target, for which GPTs supply an answer. For instance, regarding the 2016 presidential election, the prompt inquired, "Interviewer: Which candidate, did you vote for in the 2016 presidential election? Did you vote for 'Hillary Clinton (Democrat)' or 'Donald Trump (Republican)'?" For global warming beliefs with the binary answer option, it was phrased as, "Interviewer: What do you think: Do you think that global warming is happening? Would you say 'Yes', or 'No'?" For other target questions about global warming, we provided comprehensive answer options that matched the survey. For global warming belief with multiple response options, the target question was phrased as, "Interviewer: What do you think: Do you think that global warming is happening? Would you say 'Yes', 'Don't know', 'No', or 'Refused' to answer?" For the causation of global warming, the target question was phrased as, "Interviewer: Assuming global warming is happening, do you think it is 'Caused mostly by natural changes in the environment', 'Caused mostly by human activities', 'Caused by both human activities and natural changes', 'Neither because global warming isn't happening', 'Other (Please specify)', 'Don't know', or 'Refused' to answer?" For global warming worry, the target question was, phrased as "Interviewer: How worried are you about global warming? Would you say you are 'Not at all worried', 'Not very worried', 'Somewhat worried', 'Very worried', or 'Refused' to answer?"

Occasionally, GPTs generated answers that did not precisely match the listed options. We manually corrected these hallucinations. For instance, instead of a straightforward 'Yes,' GPT might produce, 'Yes, I believe global warming is happening.' Such responses were recoded to align with the intended options. Any deviations were easily identifiable and adjusted to fit within the given answer choices.

## 2.3. Survey Measurements

*Target variables*

*Presidential election voting behaviors*. In the 2017 survey, presidential election voting behavior was measured with the question, "Which candidate, if any, did you vote for in the 2016 presidential election?" Respondents were given five answer options: "Hillary Clinton (Democrat)," "Donald Trump (Republican)," "Another candidate," "Did not vote for any candidate for president," and "Refused." In the 2021 survey, the question and answer options remained consistent except for changes in the election year and the name of the Democratic candidate. Specifically, "2016" was replaced with "2020" in the question, and "Hillary Clinton (Democrat)" was replaced with "Joe Biden (Democrat)."

*Global warming belief*. To measure belief in global warming, we provided a brief definition of global warming as such "Global warming refers to the idea that the world's average temperature has been increasing over the past 150 years, may be increasing more in the future," and then asked "Do you believe that global warming is happening?" with three response options: "No," "Don't know," and "Yes."

*Global warming cause*. We used a recoded version of the survey question. Originally, the survey asked: "Assuming global warming is happening, do you think it is…" with five answer options: "Caused mostly by human activities," "Caused mostly by natural changes in the environment," "None of the above because global warming isn't happening," "Other (Please specify)," "Refused." This measure was then recoded to incorporate open-ended responses, expanding the original five answer choices to seven categories: "Caused mostly by human activities," "Caused mostly by natural changes in the environment," "Caused by human activities and natural changes," "Neither because global warming isn't happening," "Don't know," "Other (Please specify)," and "Refused." This recoded version was used in the LLM prompt.

*Global warming worry*. This was measured with a question asking "How worried are you about global warming?" with four response options: "Not at all worried," "Not very worried," "Somewhat worried," and "Very worried."

### *Demographics*

Demographic details such as race, ethnicity, gender, age, education, and residential state were provided by Ipsos, based on answers provided when enrolling panel members. Race and ethnicity used five categories: "White, Non-Hispanic," "Black, Non-Hispanic," "Other, Non-Hispanic," "Hispanic," and "2+ Races, Non-Hispanic." Gender included two categories: "Male," and "Female." Age was segmented into four groups: "18-29," "30-44," "45-59," and "66+." Education was segmented into four categories: "Less than high school," "High school," "Some college," "Bachelor's degree or higher." Residential state includes 50 states and the District of Columbia of the U.S.

*Political ideology*. This was measured with a question asking "In general, do you think of yourself as…" with six response options: "Very liberal," "Somewhat liberal,", "Moderate, middle of the road," "Somewhat conservative," "Very conservative."

*Political party*. We employed a two-step method to gauge political party. First, participants were asked to identify themselves as "Republican," "Democrat," "Independent," "Other," or "No party/Not interested in politics." Those who chose "Independent" or "Other" were then asked a second question: whether they were more aligned with the "Republican party," "Democratic party," or "Neither." If participants initially identified as Republican or Democrat, or if they leaned towards one of these parties in the secondary question, they were categorized accordingly. Those who answered "Independent" in the first question or "Neither" in the second question were categorized into "Independent/Other." Participants who responded with "No party/Not interested in politics" were categorized into "No party/Not interested."

***Covariates***

*Issue involvement in global warming.* This was measured with a question asking "How important is the issue of global warming to you personally?" with six response options: 'Not at all important', 'Not too important', 'Somewhat important', 'Very important', 'Extremely important' and 'Refused.'

*Interpersonal discussion about global warming.* This was measured with a question asking 'How often do you discuss global warming with your family and friends?' with five response options: 'Never', 'Rarely', 'Occasionally', 'Often,' and "Refused."

*Awareness of scientific consensus.* This was measured with a question asking 'Which comes closest to your own view?' with five response options: 'Most scientists think global warming is not happening', 'There is a lot of disagreement among scientists about whether or not global warming is happening', 'Most scientists think global warming is happening', 'Don't know enough to say' and "Refused."

**3. Results**

**3.1. Algorithmic Fidelity of Presidential Election: A Replication**

First, we replicated a previous study (Argyle et al., 2023) which aimed to assess algorithmic fidelity in predicting voting behaviors during presidential elections. The original study concentrated on binary outcomes, examining votes for either Democratic or Republican candidates. Mirroring this approach, we narrowed our sample to participants who voted for one of these two presidential candidates. Consequently, GPT models were restricted to this binary choice when assessing voting behaviors. Our findings align with the earlier research, confirming that LLMs can effectively mimic voting behaviors when adjusted for individual demographics. However, an exception emerged with GPT-3.5, which displayed inaccuracy in predicting the voting choices of individuals who supported the Republican candidate in the 2020 presidential election. When this outlier is excluded, the average accuracy stands at 91% across all models and years ($SD$ = 1.53), signifying a substantial level of algorithmic fidelity in predicting voting behaviors in presidential elections. Figure 1 illustrates the rate at which GPTs accurately predicted voting behaviors for each candidate in the 2016 and 202 presidential elections, drawing from survey data gathered in 2017 and 2021, respectively.

*Figure 1 about here*

**3.2. Algorithmic Fidelity of Global Warming Belief: From Binary Choice to Polynomial Choice**

Can LLMs demonstrate a similar high level of algorithmic fidelity for beliefs in global warming as they did in voting behaviors? For a fair comparison between voting behaviors and global warming belief, we limited our sample to respondents who answered either "Yes" or "No" that global warming is happening. Similarly, GPT models were constrained to these binary responses. The average accuracy of GPTs across the models, conditions, and years was 85% ($SD$ = 3.41), which suggests that GPTs predict the belief that global warming is happening with a high accuracy.

Accuracy, while an intuitive measure of correct predictions, can be misleading in datasets with skewed distributions, such as our data on belief that global warming is happening. For example, if a majority of survey participants respond with "Yes" and GPTs predict all cases as "Yes", they can still appear highly accurate. A F1 score that accounts for both precision and recall offers a balanced evaluation when dealing with data that is unevenly distributed. To get a more nuanced understanding, we assessed the models using the F1 score and the Macro-Average F1 score (MAF1) which averages F1 scores across labels.

When GPTs were only conditioned on demographics, their prediction was notably compromised. These models displayed high F1 scores for "Yes" predictions (F1 range: .91-.92), but the F1 scores for "No" predictions were significantly low or even missing due to the scarcity of "No" outcomes in the silicon samples (F1 range: NA-.08). Surprisingly, these models seem to assume a universal belief in global warming, an assumption that does not accurately reflect the diversity of real-world viewpoints. To enhance the algorithmic fidelity of LLMs, it is crucial to introduce additional covariates relevant to global warming, such as issue involvement, intentions for interpersonal communication about global warming, and perceived scientific consensus about global warming. When GPT-4 was conditioned on both demographics and these additional covariates, its MAF1 improved from .49 to .82 in the 2017 survey and from unavailable to .85 in 2021. Similarly, under the same conditionings, GPT-3.5's MAF1 increased from unavailable to .53 in 2017 and to .65 in 2021.

In our previous experiment, GPTs were limited to a binary choice when expressing its response to beliefs that global warming is happening. But what happens if we introduce a third option, "Don't know," as is typically done in surveys? It turns out that the introduction of an additional response option decreased the accuracy of both GPTs. The average accuracy of GPTs across the models, conditions, and years was 75% ($SD = 3.70$), which is lower than the accuracy of GPTs with binary choices. Notably, GPTs that were solely conditioned on demographics did not generate "No" or "Don't know" responses at all, resulting in F1 scores for these responses as non-existent. GPTs that were conditioned on demographics and covariates improved their performances. Nevertheless, an examination of the F1 score indicates that GPT-4 had more difficulty predicting "Don't Know" (F1 in 2017: .16; F1 in 2021: .20) than "No" (F1 in 2017: .58; F1 in 2021: .60), while GPT-3.5 struggled with predicting both "No" (F1 in 2017: .24; F1 in 2021: .21) and "Don't know" (F1 in 2017: .32; F1 in 2021: .34).

Figure 2 displays the response distributions from both survey participants and silicon samples regarding their belief that global warming is happening, with two (upper panel) and three response choices (lower panel). Similar to the binary version, models that were conditioned solely on demographics with three answer choices overestimated the proportion of individuals who believe global warming is happening, compared to the actual survey results. When GPTs were conditioned on both demographics and covariates, response distributions aligned more with the survey data.

*Figure 2 about here*

### 3.3. Algorithmic Fidelity of Global Warming Cause

We then asked GPT models about the cause of global warming, which we then compared to the responses of survey participants. The survey asked respondents to choose from a predefined list of options regarding the causes of global warming, denoted as "Human", "Nature", "Both human and nature", "Global warming isn't happening", "Other", "Don't know", and "Refused." (The labels have been rephrased for simplicity in the text. The full verbatim is available in the Methods section). We faithfully replicated these response options in our prompts. The average accuracy of GPTs across the models, conditions, and years was 51% ($SD = 7.42$). MAF1s were not available for these models, as each failed to produce F1 scores for certain answer options. When F1 scores were missing, we did not compute MAF1.

Figure 3 provides an overview of the response distributions from both the survey and the silicon samples, pertaining to the causation of global warming. Interestingly, the GPT-4 model, when solely conditioned on demographics, significantly overestimates the proportion of individuals who attribute global warming primarily to human activities than the GPT-3.5 model under the same conditioning. This result came as a surprise, especially given GPT-4's overall superiority over GPT-3.5 in various cognitive tasks. However, conditioning models on demographics and covariates tend to make the distribution of responses more aligned with that of the survey.

*Figure 3 about here*

### 3.4. Algorithmic Fidelity of Global Warming Worry: From Categorical Answers to Ordinal Assessment

In the final phase, we asked the GPT models about their estimated level of worry about global warming. In the survey, this question was structured as an ordinal variable with four distinct categories: "Very Worried," "Somewhat Worried," "Not Very Worried," and "Not At All Worried." We incorporated this ordinal scale in the prompts provided to the GPT models. The average accuracy of GPTs across the models, conditions, and years was 48% ($SD = 13.02$). When GPT-4 was conditioned solely on demographics, its predictions did not reflect the survey data (MAF1 in 2017 = .22, MAF1 in 2021 = .22). Moreover, GPT-3.5 with demographics only did not produce any "Not very worried" and "Not at all worried" responses, making MAF1 unavailable for both years. Conditioning the models to demographics and additional covariates improved the algorithmic fidelity, with GPT-4 exhibiting superior performance (MAF1 in 2017 = .65, MAF1 in 2021 = .54) over GPT-3.5 (MAF1 in 2017 = .47, MAF1 in 2021 = .50).

The interplay between conditions and model version that affects algorithmic fidelity is reflected in Figure 4. Both GPT-4 and GPT-3.5, conditioned on demographics alone, overestimated the proportion of individuals who were either "very worried" or "somewhat worried" about global warming. Notably, similar to the earlier findings, GPT-4, when conditioned solely on demographics, displayed more extreme estimations than GPT-3.5, particularly overestimating the percentage of individuals who expressed being "very worried" about global warming. On the other hand, GPT-4 conditioned on both demographics and covariates displayed a response distribution that aligned more with the survey data than GPT-3.5 that had the equivalent conditions.

*Figure 4 about here*

### 3.5. Assessment of Overall Distributions between Survey Samples and Silicon Samples

Figure 5 illustrates the divergence in response patterns between survey samples and silicon samples. The smaller the dot size, the closer the distribution pattern is to the survey data. Across the global warming variables, GPTs that are conditioned on both demographics and covariates display less deviation from the survey data. This suggests that these models produce response patterns more in line with the survey data. Notably, under these conditions, GPT-4's response distribution tends to be more consistent with the survey data than those of GPT-3.5.

*Figure 5 about here*

### 3.6. Algorithmic Bias Assessment across Sub-populations

Here we examine how well GPT models represent presidential voting behaviors and belief that global warming is happening across various sub-populations. We focus on the models with high fidelity, specifically GPT-4 conditioned solely on demographics predicting voting behaviors, and GPT-4 conditioned on both demographics and covariates predicting the binary version of belief that global warming is happening. In Table 1, we present accuracy and MAF1 results. We based our interpretations on the MAF1, where a score below 0.70 conventionally indicates inadequate performance.

The overarching findings indicate that GPT-4 accurately predicted voting behaviors and belief that global warming is happening among diverse sub-populations. However, certain sub-populations were less accurate. Notably, GPT-4 fell short in accurately predicting the voting behavior of Non-Hispanic Blacks in the presidential election in the 2017 survey (MAF1 = .61) and their belief that global warming is happening in both years (MAF1 in 2017 = .62, MAF1 in 2021 = .60). Further analysis reveals that GPT-4 underestimated the proportion of Non-Hispanic Blacks who voted for the Democratic candidate in 2016 and those who believed global warming is happening. Additionally, GPT-4 also underrepresented the belief that global warming is happening among Non-Hispanic Others in 2021 (MAF1 = .64), although this requires further investigation due to the limited sample size for this subgroup and the isolated nature of the result.

In 2017, GPT-4 similarly underestimated Democrats who believed in global warming, though its predictions improved in 2021. Regarding the prediction of voting behaviors based on political party affiliation, GPT-4 yielded subpar F1 scores. This is attributed to GPT-4 overestimating voting behaviors based on an individual's political affiliation. For instance, while 95% of Democrats reported voting for Hillary Clinton in the 2017 survey, GPT-4 predicted that 99% of Democrats voted for the candidate. Similarly, with 94% of Republicans reporting voting for Donald Trump, GPT-4 predicted that 100% of Republicans voted for the Republican candidate. Such overestimations render certain F1 scores unattainable when considering political parties and ideologies.

*Table 1 about here*

### 3.7. Pattern Correspondence Assessment

It is important that GPT-generated responses "reflect underlying patterns of relationships between ideas, demographics, and behavior that would be observed in comparable human-produced data" (Argyle et la., 2023, p. 340). We investigated the extent to which the outputs from the GPTs reflect the correlations between demographics and each variable found in the survey data. We used Cramer's V to measure the strength of association between demographics and each of the target variables. Figure 6 illustrates the values of Cramer's V between demographics and each variable for both survey and GPT models. Notably, GPT-4, when conditioned on both demographics and covariates, exhibits the closest correspondence to the survey data in terms of association patterns (Cramer's V mean difference between survey and the model (hereafter, *diff*) = .03, *SD* = .03). This is followed by GPT-3.5 conditioned on demographics and covariates (*diff* = .06, *SD* = .08), GPT-4 conditioned on demographics only (*diff* = .07, *SD* = .07), and GPT-3.5 conditioned on demographics only (*diff* = .08, *SD* = .10).

*Figure 6 about here*

## 4. Discussion

This research investigates the algorithmic fidelity and bias of Large Language Models (LLMs) by simulating public opinion about global warming, and comparing the synthesized data with survey data. Initially, we replicate the study conducted by Argyle et al. (2023) using a novel dataset. In line with previous research, our results indicate that LLMs are generally adept at replicating presidential voting behaviors. We extend our analysis to the realm of global warming, revealing that LLMs exhibit promising capabilities in predicting global warming opinions. Nevertheless, our findings identify several concerns that scholars must consider when employing LLMs in global warming survey research.

Our empirical findings underscore the importance of including relevant psychological covariates, in addition to demographics, to achieve a high level of algorithmic fidelity in global warming research. In our analysis, LLMs conditioned on both demographics and issue-related covariates show a significant improvement in predicting individual beliefs in global warming compared to LLMs using demographics alone. This finding demonstrates LLMs' promising capability that accounts for psychological covariates and prioritizes these more influential factors over demographics in predicting global warming perceptions. This aligns with findings from climate change research demonstrating psychological factors are more strongly associated with beliefs in global warming than demographics (Hornsey et al., 2016). However, it is also worth noting that LLMs, when conditioned solely on demographics, fail to adequately integrate demographic factors such as age, education, political affiliation, and ideology known to correlate with global warming belief to some degree (Hornsey et al., 2016), leading to unrealistic predictions where everyone believes in global warming. This phenomenon is consistent across various global warming scenarios, painting an inaccurate picture of public beliefs and worry about global warming.

The LLM version also impacts algorithmic fidelity. When LLMs are conditioned solely on demographics, GPT-4 tends to produce more extreme results than GPT-3.5. For example, GPT-4, when conditioned solely on demographics, significantly overestimates the proportion of individuals who believe that global warming is caused by human activities and those who express high levels of global warming worry (i.e., "very worried"). However, this disparity is considerably reduced when LLMs are conditioned with additional issue-related covariates. GPT-4, when conditioned with demographics and covariates, provides more accurate predictions regarding public perceptions of global warming and displays response distributions more closely aligned with survey results. This suggests that the extended "training" and "alignment" procedures employed by GPT-4 may impact algorithmic fidelity both positively and negatively, highlighting a nuanced trade-off between model complexity and fidelity, particularly in the context of science-based subjects like global warming. The opacity of AI corporations like OpenAI in the development of LLMs makes it difficult to discern the factors that influence LLMs' responses, especially when faced with uncertainty in individual survey responses (i.e., the demographics only condition). It raises questions about whether massive training data favor certain attitudes toward climate change, or if the post-training adjustments guided by human feedback lack representativeness. To enhance the reliability of LLMs for social science research, transparency in LLM development is imperative.

Our data reveal algorithmic bias regarding specific sub-populations. Prior research has also highlighted such bias, particularly in LLMs refined by human feedback (Santurkar et al., 2023). The estimated opinions from these LLMs tend to align more with individuals who are liberal, have higher incomes, higher education, and those who identify as non-religious or follow religious faiths other than Buddhism, Islam, and Hinduism (Santurkar et al., 2023). In our study, LLMs struggle to accurately predict the voting behaviors and beliefs about global warming of Black Americans. This finding cannot be solely attributed to the sample size, as evaluation metrics are still adequate for other racial and multi-racial groups with even smaller sample sizes. Such inaccuracies can lead to skewed predictions and analyses, potentially marginalizing these groups and disregarding their specific concerns, particularly in discussions on critical social issues like climate justice. A comprehensive examination of algorithmic bias in LLMs regarding marginalized groups is crucial, not only in the context of algorithmic fidelity but also in various other applications of LLMs in social science research.

Our research offers practical guidance for using LLMs in global warming survey research and beyond. First, climate change researchers should condition LLMs based on domain knowledge and scientific evidence. Including covariates known to correlate with perceptions of global warming, such as issue involvement, interpersonal communication engagement, and scientific consensus, can improve algorithmic fidelity. As perceptions of global warming are multifaceted, the inclusion of other relevant covariates may further enhance the fidelity of LLMs for the researcher's target variable. Second, it is advisable to utilize advanced models like GPT-4. When LLMs are conditioned with relevant covariates, the algorithmic fidelity of GPT-4 surpasses that of GPT-3.5 in all scenarios. Third, researchers should consider minimizing the number of answer options, especially for indeterminate answers like "Don't Know," "Refused," and "Other." As with other prediction models, the predictability of LLMs diminishes as the number of answer options increases. Inclusion of indeterminate answers reduces the algorithmic fidelity of LLMs more significantly. While limiting answer options might diminish the utility of LLMs in survey research, it still holds significant value in streamlining exploratory data collection for social scientists. For instance, our study collected responses from 1304 silicon samples at a cost of approximately $2.08 using GPT-3.5 and $20.86 using GPT-4, showcasing marked cost and time savings compared to conventional data collection methods. These significantly cheaper synthesized responses could be invaluable for designing survey research and projecting survey outcomes.

**Limitations and Future Research Directions**

Our research is not without limitations, which point towards future research directions. First, our study primarily focuses on closed-ended questions, neglecting the richer insights provided by open-ended questions. Answers to open-ended questions can offer more qualitative perspectives, beliefs, and opinions, potentially enhancing LLM algorithmic fidelity. Future research should explore how incorporating responses from open-ended questions as conditional inputs influences LLM fidelity. Second, our research focuses on a narrow aspect concerning the impact of prompt format (e.g., the number of answer options) on algorithmic fidelity, leaving unexplored other facets of prompt structure that may affect fidelity. Previous studies have noted that prompt formats, such as the order of answer options in conditional inputs, influence algorithmic fidelity (Pezeshkpour & Hruschka, 2023). Beyond the order of answer options, structural factors like the order of questions, or the number and order of target questions may impact algorithmic fidelity. Investigating how different survey formats influence responses from both humans and silicon samples would be intriguing, especially given the meticulous attention survey researchers pay to the order of survey questions, which can interact with one another and elicit varying responses (McFarland, 1981). This raises the question of whether such factors matter to LLMs as much as they do to humans.

## 5. Conclusion

This research provides valuable insights into the algorithmic fidelity and bias of LLMs in simulating public opinions regarding presidential elections and global warming. Our findings indicate that LLMs can effectively replicate voting behaviors when conditioned solely on demographics. However, for complex societal issues like global warming, the inclusion of relevant issue-related covariates is essential to enhance algorithmic fidelity. When appropriately conditioned, the advanced GPT-4 model outperforms GPT-3.5, suggesting that increased model complexity can improve fidelity. Our results also shed light on gaps and biases in LLM representations, particularly for marginalized groups like Black Americans, whose voting behaviors and climate beliefs are not adequately captured compared to other racial groups. This highlights the need for comprehensive assessments of algorithmic bias across various tasks and populations. Questions around the impact of prompt structure on fidelity warrant further investigation. In conclusion, this study offers practical guidance on conditioning prompts and selecting models to maximize fidelity in social science applications while emphasizing the importance of validating LLMs, particularly for minority groups. A nuanced approach is required to harness the power of LLMs while addressing their limitations through proactive algorithm auditing and bias mitigation.

Table 1. Accuracy and MAF1 of GPTs for Presidential Election and Global Warming Belief across Sub-populations

| Variables | Election 2017 GPT4 Demo | | Election 2021 GPT4 Demo | | GW Belief 2017 GPT4 Demo + Cov | | GW Belief 2021 GPT4 Demo + Cov | |
|---|---|---|---|---|---|---|---|---|
| | Acc | MAF1 | Acc | MAF1 | Acc | MAF1 | Acc | MAF1 |
| Race/ethnicity | | | | | | | | |
| 2+ Races, Non-Hispanic<br>2017 EL $n = 20$<br>2021 EL $n = 18$<br>2017 GW $n = 32$<br>2021 GW $n = 18$ | .85 | .85 | 1.00 | 1.00 | .94 | .86 | 1.00 | 1.00 |
| Black, Non-Hispanic<br>2017 EL $n = 89$<br>2021 EL $n = 67$<br>2017 GW $n = 104$<br>2021 GW $n = 73$ | .92 | .61 | .93 | .75 | .90 | .62 | .92 | .60 |
| Hispanic<br>2017 EL $n = 78$<br>2021 EL $n = 74$<br>2017 GW $n = 132$<br>2021 GW $n = 101$ | .96 | .95 | .84 | .80 | .90 | .77 | .92 | .82 |
| Other, Non-Hispanic<br>2017 EL $n = 30$<br>2021 EL $n = 31$<br>2017 GW $n = 51$<br>2021 GW $n = 38$ | 1.00 | 1.00 | .94 | .92 | .96 | .91 | .89 | .64 |
| White, Non-Hispanic<br>2017 EL $n = 668$<br>2021 EL $n = 614$<br>2017 GW $n = 786$<br>2021 GW $n = 669$ | .91 | .91 | .90 | .90 | .88 | .82 | .90 | .86 |
| Gender | | | | | | | | |
|   Female<br>  2017 EL *n* = 454<br>  2021 EL *n* = 403<br>  2017 GW *n* = 568<br>  2021 GW *n* = 454 | .93 | .93 | .89 | .89 | .89 | .80 | .92 | .85 |
|   Male<br>  2017 EL *n* = 431<br>  2021 EL *n* = 401<br>  2017 GW *n* = 537<br>  2021 GW *n* = 445 | .91 | .91 | .90 | .90 | .89 | .84 | .90 | .85 |
| Age | | | | | | | | |
|   18-29<br>  2017 EL *n* = 71<br>  2021 EL *n* = 84<br>  2017 GW *n* = 145<br>  2021 GW *n* = 113 | .92 | .91 | .95 | .95 | .91 | .85 | .89 | .72 |
|   30-44<br>  2017 EL *n* = 172<br>  2021 EL *n* = 138<br>  2017 GW *n* = 243<br>  2021 GW *n* = 171 | .92 | .91 | .93 | .91 | .88 | .78 | .94 | .87 |
|   45-59<br>  2017 EL *n* = 242<br>  2021 EL *n* = 214<br>  2017 GW *n* = 299<br>  2021 GW *n* = 230 | .90 | .91 | .85 | .85 | .88 | .78 | .89 | .84 |
|   60+<br>  2017 EL *n* = 400<br>  2021 EL *n* = 368<br>  2017 GW *n* = 418<br>  2021 GW *n* = 385 | .92 | .92 | .90 | .90 | .89 | .85 | .91 | .87 |
| Political ideology | | | | | | | | |
|   Moderate, middle of the road<br>  2017 EL *n* = 300<br>  2021 EL *n* = 303<br>  2017 GW *n* = 424<br>  2021 GW *n* = 380 | .85 | .85 | .82 | .80 | .90 | .74 | .93 | .82 |

| Group | | | | | | | | |
|---|---|---|---|---|---|---|---|---|
| Somewhat conservative<br>2017 EL *n* = 193<br>2021 EL *n* = 175<br>2017 GW *n* = 206<br>2021 GW *n* = 165 | .94 | .89 | .90 | .81 | .81 | .80 | .81 | .80 |
| Somewhat liberal<br>2017 EL *n* = 201<br>2021 EL *n* = 157<br>2017 GW *n* = 262<br>2021 GW *n* = 168 | .94 | .77 | .97 | .82 | .96 | .68 | .98 | .83 |
| Very conservative<br>2017 EL *n* = 99<br>2021 EL *n* = 87<br>2017 GW *n* = 99<br>2021 GW *n* = 84 | .97 | .86 | .94 | .76 | .76 | .72 | .83 | .82 |
| Very liberal<br>2017 EL *n* = 90<br>2021 EL *n* = 79<br>2017 GW *n* = 104<br>2021 GW *n* = 88 | .97 | .90 | .99 | - | .97 | .78 | .98 | - |
| Political party | | | | | | | | |
| Democrats<br>2017 EL *n* = 442<br>2021 EL *n* = 397<br>2017 GW *n* = 535<br>2021 GW *n* = 423 | .95 | .53 | .96 | - | .96 | .62 | .97 | .74 |
| Independent/Other<br>2017 EL *n* = 63<br>2021 EL *n* = 63<br>2017 GW *n* = 108<br>2021 GW *n* = 96 | .68 | .65 | .59 | .40 | .89 | .84 | .91 | .84 |
| No party/Not interested<br>2017 EL *n* = 27<br>2021 EL *n* = 8<br>2017 GW *n* = 105<br>2021 GW *n* = 50 | .59 | .66 | .38 | - | .81 | .73 | .84 | .78 |
| Republicans<br>2017 EL *n* = 353 | .94 | - | .89 | - | .81 | .81 | .83 | .83 |
| 2021 EL *n* = 334
2017 GW *n* = 354
2021 GW *n* = 321 | | | | | | | | |
| Education | | | | | | | | |
| Bachelor's degree or higher
2017 EL *n* = 371
2021 EL *n* = 331
2017 GW *n* = 440
2021 GW *n* = 358 | .95 | .94 | .91 | .90 | .92 | .84 | .93 | .86 |
| High school
2017 EL *n* = 213
2021 EL *n* = 195
2017 GW *n* = 272
2021 GW *n* = 226 | .88 | .89 | .86 | .86 | .86 | .81 | .88 | .83 |
| Less than high school
2017 EL *n* = 34
2021 EL *n* = 25
2017 GW *n* = 64
2021 GW *n* = 42 | .85 | .87 | .88 | .88 | .88 | .74 | .98 | .96 |
| Some college
2017 EL *n* = 267
2021 EL *n* = 253
2017 GW *n* = 329
2021 GW *n* = 273 | .91 | .91 | .91 | .91 | .88 | .82 | .89 | .84 |

*Note*: EL represents presidential election and GW represents global warming belief. The numbers in the variable column indicate the number of sub-populations used in the analysis.

Figure 1. Proportion of Accurate Predictions by GPTs for Each Candidate in Presidential Elections

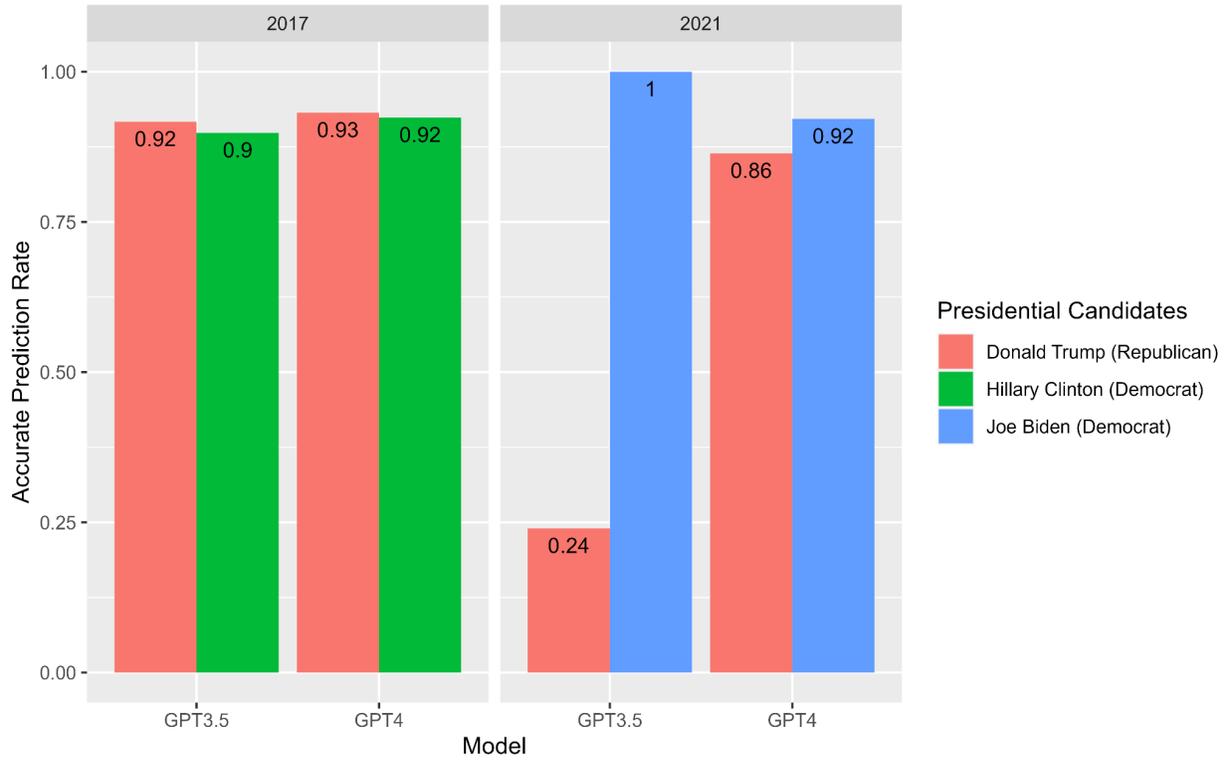

*Note*: 2016 and 2021 presidential election voting behaviors are asked in 2017 and 2021 surveys, respectively. GPT-3.5 and GPT-4 are conditioned to demographics in this context.

Figure 2. Belief that Global Warming is Happening: Distributional Comparison of Survey and Silicon Samples

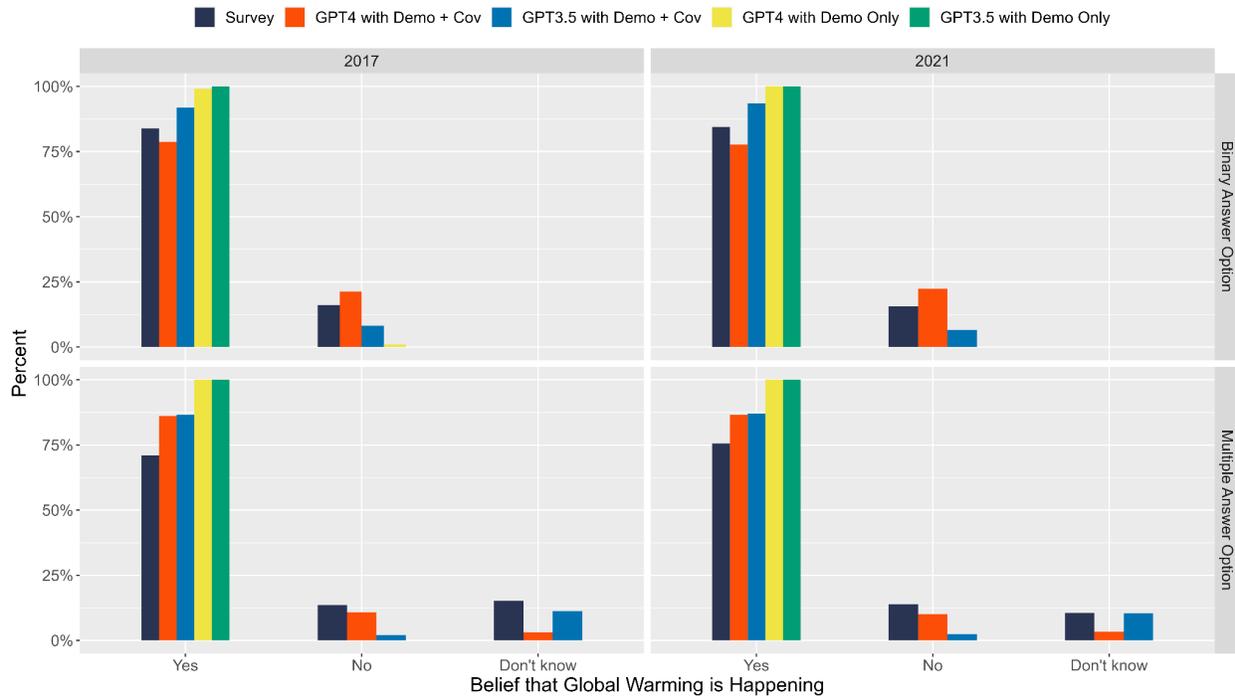

*Note*: "Demo Only" represents GPTs are conditioned solely on demographics and "Demo + Cov" represents GPTs are conditioned on demographics and covariates.

Figure 3. Global Warming Cause: Distributional Comparison of Survey and Silicon Samples

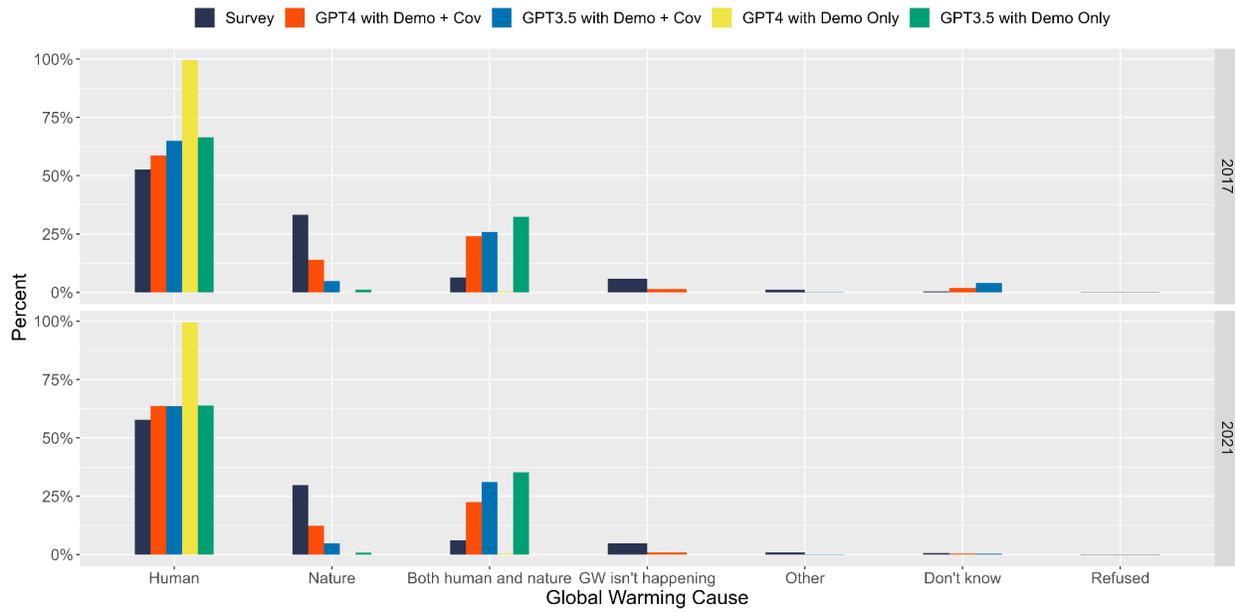

*Note*: "Demo Only" represents GPTs are conditioned solely on demographics and "Demo + Cov" represents GPTs are conditioned on demographics and covariates.

Figure 4. Global Warming Worry: Distributional Comparison of Survey and Silicon Samples

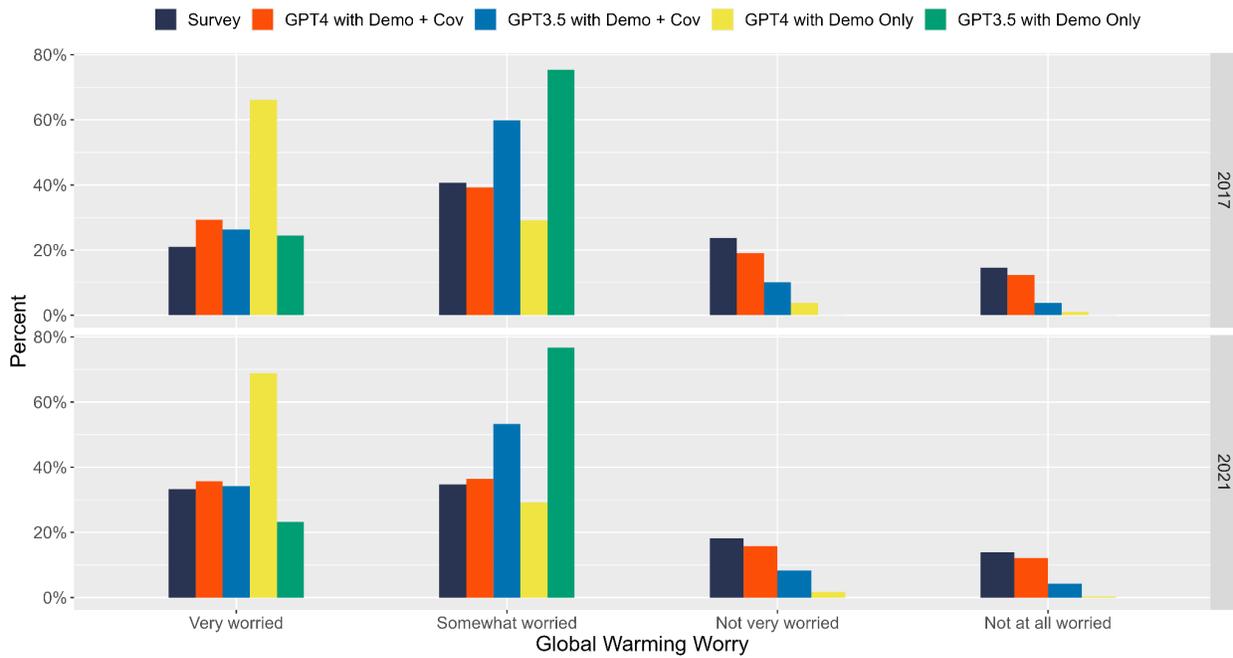

*Note*: "Demo Only" represents GPTs are conditioned solely on demographics and "Demo + Cov" represents GPTs are conditioned on demographics and covariates.

Figure 5. Kullback-Leibler Divergence (KLD) between GPTs and Survey Data.

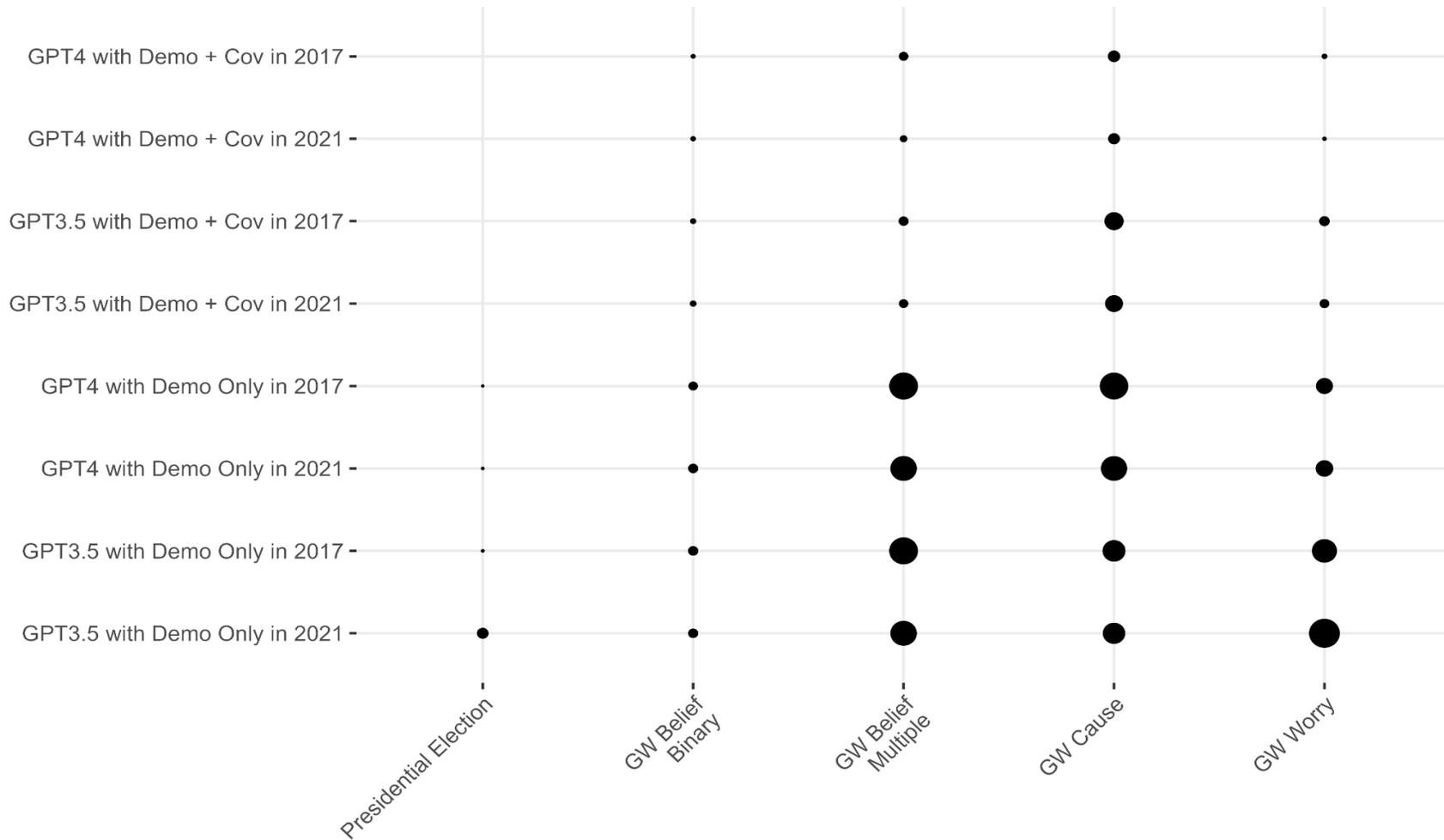

*Note*: Kullback-Leibler Divergence (KLD) is employed to evaluate the similarity in response distributions between GPT-generated responses and survey data. KLD values can vary from 0, signaling identical distributions, to infinity. In our study, KLD values fall within the range of 0.0003 to 4.26. Point sizes are adjusted proportionately based on KLD values, with smaller points indicating a greater resemblance between the distributions of GPT-generated responses and survey data. "GW" refers to Global Warming.

Figure 6. Cramer's V correlations in Survey vs. GPTs

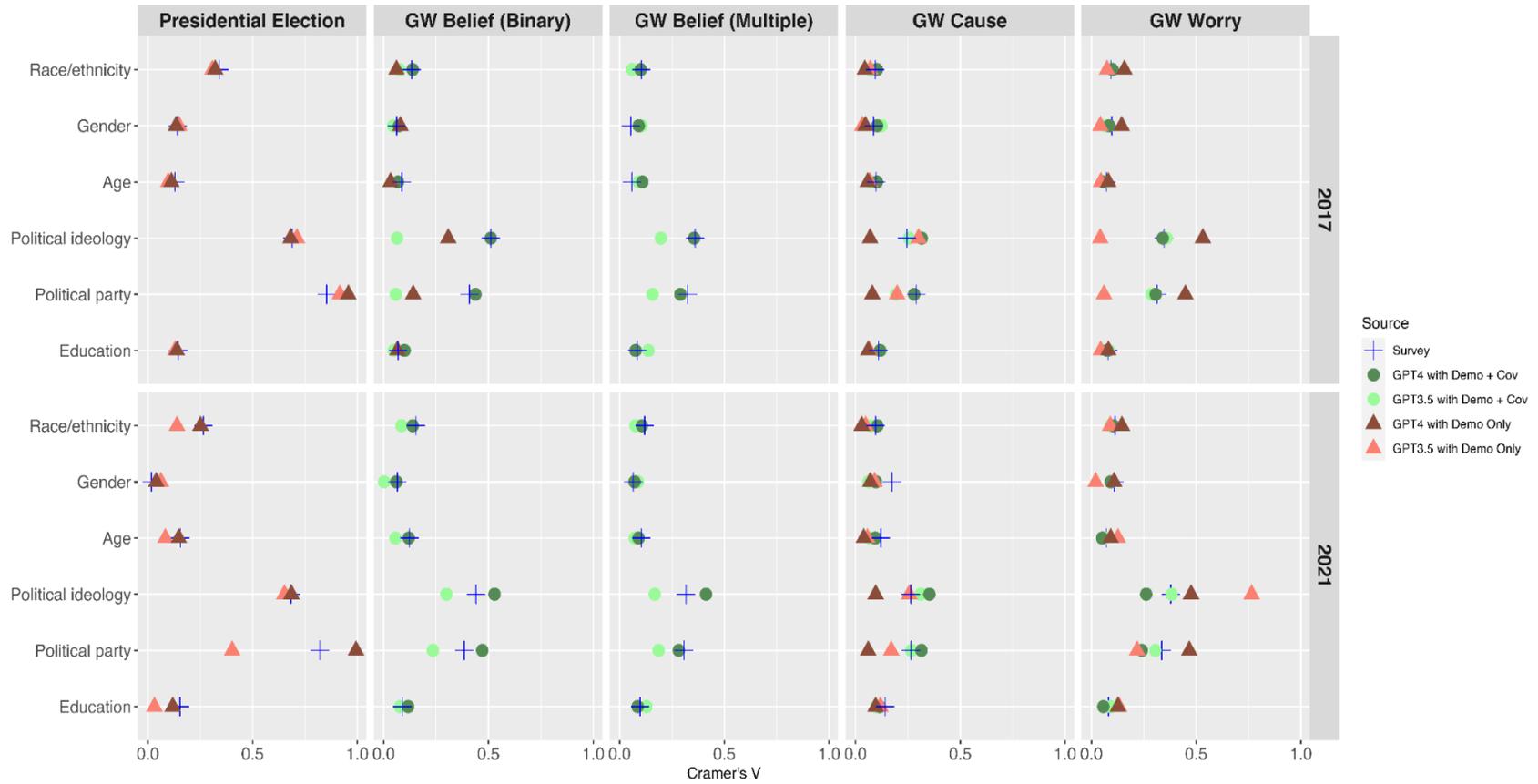

Note: GPT4 with Demo + Cov and GPT3.5 with Demo + Cov are absent in the analysis of the presidential election. Moreover, Cramer's V could not be estimated for some models due to no variation in the outcome (e.g., GPT3.5 with Demo Only for GW belief)

Supplemental Table 1. Prompt Examples for Models Conditioned Solely on Demographics

> Role: System
>
> Content:
>
> You are an interviewee. Based on your previous answers, provide an answer to the last question
>
> Role: User
>
> Content:
>
> Interviewer: What is the current year and month in which this interview is being conducted?
>
> Me: October 2017
>
> Interviewer: I am going to read you a list of five race categories. What race do you consider yourself to be? 'White, Non-Hispanic', 'Black, Non-Hispanic', '2+ Races, Non-Hispanic', 'Hispanic', or 'Other, Non-Hispanic'
>
> Me: {race}
>
> Interviewer: What is your gender? Do you identify as 'Male' or 'Female'?
>
> Me: {gender}
>
> Interviewer: What age category does your age fall in? '60+','45-59','30-44', or '18-29'
>
> Me: {age}
>
> Interviewer: Which would you say best describes your political ideology. Would you say you are a 'Very conservative','Somewhat conservative','Moderate, middle of the road','Somewhat liberal','Very liberal',or 'Refused' to answer?
>
> Me: {ideology}
>
> Interviewer: Which would you say best describes your partisan identification. Would you say you are a 'Republicans', 'Democrats', 'No party/Not interested', 'Independent/Other', or 'Refused'to answer?
>
> Me: {party}
>
> Interviewer: What is the highest level of school you have completed, or the highest degree you have received? Is it 'Bachelor's degree or higher','High school', 'Some college', or 'Less than high school'?
>
> Me: {education}
>
> Interviewer: In which state do you reside?
>
> Me: {state}

> Interviewer: What do you think: Do you think that global warming is happening? Would you say 'Yes','dont know', 'No', or 'Refused' to answer?
>
> Me:

*Note*: Content in curly brockets provided based on survey data. The last question is the target question that a silicon sample provides an answer.

Supplemental Table 2. Prompt Examples for Models Conditioned both on Demographics and Covariates

> Role: System
>
> Content:
>
> You are an interviewee. Based on your previous answers, provide an answer to the last question
>
> Role: User
>
> Content:
>
> Interviewer: What is the current year and month in which this interview is being conducted?
>
> Me: October 2017
>
> Interviewer: I am going to read you a list of five race categories. What race do you consider yourself to be? 'White, Non-Hispanic', 'Black, Non-Hispanic', '2+ Races, Non-Hispanic', 'Hispanic', or 'Other, Non-Hispanic'
>
> Me: {race}
>
> Interviewer: What is your gender? Do you identify as 'Male' or 'Female'?
>
> Me: {gender}
>
> Interviewer: What age category does your age fall in? '60+','45-59','30-44', or '18-29'
>
> Me: {age}
>
> Interviewer: Which would you say best describes your political ideology. Would you say you are a 'Very conservative','Somewhat conservative','Moderate, middle of the road','Somewhat liberal','Very liberal',or 'Refused' to answer?
>
> Me: {ideology}
>
> Interviewer: Which would you say best describes your partisan identification. Would you say you are a 'Republicans', 'Democrats', 'No party/Not interested', 'Independent/Other', or 'Refused'to answer?
>
> Me: {party}
>
> Interviewer: What is the highest level of school you have completed, or the highest degree you have received? Is it 'Bachelor's degree or higher','High school', 'Some college', or 'Less than high school'?
>
> Me: {education}
>
> Interviewer: In which state do you reside?

> Me: {state}
>
> Interviewer: How important is the issue of global warming to you personally? Is it 'Not at all important', 'Not too important', 'Somewhat important', 'Very important', or 'Extremely important'?
>
> Me: {issue_involvement}
>
> Interviewer: How often do you discuss global warming with your family and friends? Do you discuss the subject with your family and friends 'Never', 'Rarely', 'Occasionally', or 'Often'?
>
> Me: {discussion}
>
> Interviewer: Which comes closest to your own view? 'Most scientists think global warming is not happening', 'There is a lot of disagreement among scientists about whether or not global warming is happening', 'Most scientists think global warming is happening', or 'Don't know enough to say'?
>
> Me: {science_consensus}
>
> Interviewer: What do you think: Do you think that global warming is happening? Would you say 'Yes','dont know', 'No', or 'Refused' to answer?
>
> Me:

*Note*: Content in curly brokets provided based on survey data. The last question is the target question that a silicon sample provides an answer.

Supplemental Table 3. Evaluation Metrics on Presidential Election

| Models | Accuracy | Answers | F1 | Precision | Recall |
|---|---|---|---|---|---|
| GPT 4.0 Demo only for 2017 | .93 | Trump | .92 | .91 | .93 |
| | | Clinton | .93 | .94 | .92 |
| | | Macro-average | .93 | .93 | .93 |
| GPT 4.0 Demo only for 2021 | .90 | Trump | .88 | .89 | .86 |
| | | Clinton | .91 | .90 | .92 |
| | | Macro-average | .89 | .90 | .89 |
| GPT 3.5 Demo only for 2017 | .91 | Trump | .90 | .88 | .92 |
| | | Clinton | .91 | .93 | .90 |
| | | Macro-average | .91 | .91 | .91 |
| GPT 3.5 Demo only for 2021 | .67 | Trump | .39 | 1.00 | .24 |
| | | Clinton | .78 | .64 | 1.00 |
| | | Macro-average | .58 | .82 | .62 |

Supplemental Table 4. Evaluation Metrics on Global Warming Belief with Binary Answer Option

| Models | Accuracy | Answers | F1 | Precision | Recall |
|---|---|---|---|---|---|
| GPT 4.0 Demo + Cov for 2017 | .89 | Yes | .93 | .96 | .90 |
| | | No | .70 | .62 | .81 |
| | | Macro-average | .82 | .79 | .86 |
| GPT 4.0 Demo + Cov for 2021 | .91 | Yes | .94 | .99 | .91 |
| | | No | .76 | .65 | .93 |
| | | Macro-average | .85 | .82 | .92 |
| GPT 3.5 Demo + Cov for 2017 | .80 | Yes | .88 | .85 | .93 |
| | | No | .17 | .25 | .13 |
| | | Macro-average | .53 | .55 | .53 |
| GPT 3.5 Demo + Cov for 2021 | .86 | Yes | .92 | .88 | .97 |
| | | No | .37 | .64 | .26 |
| | | Macro-average | .65 | .76 | .62 |
| GPT 4.0 Demo only for 2017 | .84 | Yes | .91 | .84 | 1.00 |
| | | No | .08 | .70 | .04 |
| | | Macro-average | .49 | .77 | .52 |
| GPT 4.0 Demo only for 2021 | .84 | Yes | .92 | .84 | 1.00 |
| | | No | - | - | 0 |
| | | Macro-average | - | - | .50 |
| GPT 3.5 Demo only for 2017 | .84 | Yes | .91 | .84 | 1.00 |
| | | No | - | - | 0 |
| | | Macro-average | - | - | .50 |
| GPT 3.5 Demo only for 2021 | .84 | Yes | .92 | .84 | 1.00 |
| | | No | - | - | 0 |
| | | Macro-average | - | - | .50 |

Supplemental Table 5. Evaluation Metrics on Global Warming Belief with Multiple Answer Options

| Models | Accuracy | Answers | F1 | Precision | Recall |
|---|---|---|---|---|---|
| GPT 4.0 Demo + Cov for 2017 | .77 | Yes | .88 | .80 | .97 |
|  |  | No | .58 | .66 | .52 |
|  |  | Don't know | .16 | .48 | .10 |
|  |  | Macro-average | .54 | .64 | .53 |
| GPT 4.0 Demo + Cov for 2021 | .82 | Yes | .91 | .85 | .97 |
|  |  | No | .60 | .71 | .51 |
|  |  | Don't know | .20 | .41 | .13 |
|  |  | Macro-average | .57 | .66 | .54 |
| GPT 3.5 Demo + Cov for 2017 | .72 | Yes | .83 | .76 | .92 |
|  |  | No | .24 | .93 | .14 |
|  |  | Don't know | .32 | .38 | .28 |
|  |  | Macro-average | .47 | .69 | .45 |
| GPT 3.5 Demo + Cov for 2021 | .75 | Yes | .86 | .80 | .93 |
|  |  | No | .21 | .68 | .12 |
|  |  | Don't know | .34 | .34 | .34 |
|  |  | Macro-average | .47 | .69 | .45 |
| GPT 4.0 Demo only for 2017 | .71 | Yes | .83 | .71 | 1.00 |
|  |  | No | - | - | 0 |
|  |  | Don't know | - | - | 0 |
|  |  | Macro-average | - | - | .33 |
| GPT 4.0 Demo | .76 | Yes | .86 | .76 | 1.00 |
| only for 2021 | | No | - | - | 0 |
| | | Don't know | - | - | 0 |
| | | Macro-average | - | - | .33 |
| GPT 3.5 Demo only for 2017 | .71 | Yes | .83 | .71 | 1.00 |
| | | No | - | - | 0 |
| | | Don't know | - | - | 0 |
| | | Macro-average | - | - | .33 |
| GPT 3.5 Demo only for 2021 | .76 | Yes | .86 | .76 | 1.00 |
| | | No | - | - | 0 |
| | | Don't know | - | - | 0 |
| | | Macro-average | - | - | .33 |

Supplemental Table 6. Evaluation Metrics on Global Warming Cause

| Models | Accuracy | Answers | F1 | Precision | Recall |
|---|---|---|---|---|---|
| GPT 4.0 Demo + Cov for 2017 | .53 | Human | .74 | .71 | .79 |
| | | Nature | .38 | .64 | .27 |
| | | Both | .09 | .06 | .22 |
| | | Neither | .27 | .72 | .17 |
| | | Don't know | .13 | .08 | .33 |
| | | Other | - | - | 0 |
| | | Refused | - | - | 0 |
| | | Macro-average | - | - | .25 |
| GPT 4.0 Demo + Cov for 2021 | .61 | Human | .83 | .79 | .87 |
| | | Nature | .41 | .70 | .29 |
| | | Both | .14 | .09 | .33 |
| | | Neither | .17 | .56 | .10 |
| | | Don't know | - | 0 | 0 |
| | | Other | - | - | 0 |
| | | Refused | - | - | 0 |
| | | Macro-average | - | - | .23 |
| GPT 3.5 Demo + Cov for 2017 | .46 | Human | .72 | .65 | .80 |
| | | Nature | .10 | .39 | .06 |
| | | Both | .09 | .05 | .22 |
| | | Neither | - | - | 0 |
| | | Don't know | - | 0 | 0 |
| | | Other | - | 0 | 0 |
| | | Refused | - | - | 0 |
| | | Macro-average | - | - | .15 |
| GPT 3.5 Demo + Cov for 2021 | .53 | Human | .79 | .16 | .83 |
| | | Nature | .16 | .57 | .09 |
| | | Both | .11 | .07 | .34 |
| | | Neither | - | - | 0 |
| | | Don't know | - | 0 | 0 |
| | | Other | - | 0 | 0 |
| | | Refused | - | - | 0 |
| | | Macro-average | - | - | .18 |
| GPT 4.0 Demo only for 2017 | .53 | Human | .69 | .53 | 1.00 |
| | | Nature | - | 0 | 0 |
| | | Both | - | 0 | 0 |
| | | Neither | - | - | 0 |
| | | Don't know | - | - | 0 |
| | | Other | - | - | 0 |
| | | Refused | - | - | 0 |
| | | Macro-average | - | - | .14 |
| GPT 4.0 Demo only for 2021 | .58 | Human | .73 | .58 | 1.00 |
| | | Nature | .01 | 1.00 | .003 |
| | | Both | - | 0 | 0 |
| | | Neither | - | - | 0 |
| | | Don't know | - | - | 0 |
| | | Other | - | - | 0 |
| | | Refused | - | - | 0 |
| | | Macro-average | - | - | .14 |
| GPT 3.5 Demo only for 2017 | .41 | Human | .63 | .57 | .71 |
| | | Nature | .04 | .50 | .02 |
| | | Both | .12 | .07 | .37 |
|  |  | Neither | - | - | 0 |
|  |  | Don't know | - | - | 0 |
|  |  | Other | - | - | 0 |
|  |  | Refused | - | - | 0 |
|  |  | Macro-average | - | - | .16 |
| GPT 3.5 Demo only for 2021 | .41 | Human | .63 | .60 | .67 |
|  |  | Nature | .04 | .75 | .02 |
|  |  | Both | .10 | .06 | .34 |
|  |  | Neither | - | - | 0 |
|  |  | Don't know | - | - | 0 |
|  |  | Other | - | - | 0 |
|  |  | Refused | - | - | 0 |
|  |  | Macro-average | - | - | .15 |

Supplemental Table 7. Evaluation Metrics on Global Warming Worry

| Models | Accuracy | Answers | F1 | Precision | Recall |
|---|---|---|---|---|---|
| GPT 4.0 Demo + Cov for 2017 | .66 | Very worried | .73 | .63 | .88 |
| | | Somewhat worried | .67 | .68 | .66 |
| | | Not very worried | .53 | .59 | .48 |
| | | Not at all worried | .66 | .72 | .61 |
| | | Macro-average | .65 | .66 | .66 |
| GPT 4.0 Demo + Cov for 2021 | .57 | Very worried | .67 | .64 | .69 |
| | | Somewhat worried | .57 | .55 | .58 |
| | | Not very worried | .42 | .45 | .39 |
| | | Not at all worried | .50 | .54 | .47 |
| | | Macro-average | .54 | .55 | .53 |
| GPT 3.5 Demo + Cov for 2017 | .55 | Very worried | .70 | .63 | .79 |
| | | Somewhat worried | .62 | .52 | .76 |
| | | Not very worried | .22 | .37 | .16 |
| | | Not at all worried | .33 | .80 | .21 |
| | | Macro-average | .47 | .58 | .48 |
| GPT 3.5 Demo + Cov for 2021 | .60 | Very worried | .81 | .80 | .82 |
|  |  | Somewhat worried | .62 | .51 | .79 |
|  |  | Not very worried | .19 | .31 | .14 |
|  |  | Not at all worried | .36 | .77 | .24 |
|  |  | Macro-average | .50 | .60 | .50 |
| GPT 4.0 Demo only for 2017 | .31 | Very worried | .46 | .30 | .95 |
|  |  | Somewhat worried | .27 | .32 | .23 |
|  |  | Not very worried | .07 | .27 | .04 |
|  |  | Not at all worried | .09 | .69 | .05 |
|  |  | Macro-average | .22 | .39 | .32 |
| GPT 4.0 Demo only for 2021 | .39 | Very worried | .61 | .45 | .93 |
|  |  | Somewhat worried | .26 | .28 | .23 |
|  |  | Not very worried | .02 | .12 | .01 |
|  |  | Not at all worried | .01 | .33 | .01 |
|  |  | Macro-average | .22 | .29 | .30 |
| GPT 3.5 Demo only for 2017 | .35 | Very worried | .19 | .18 | .20 |
|  |  | Somewhat worried | .52 | .40 | .74 |
|  |  | Not very worried | - | - | 0 |
|  |  | Not at all worried | - | - | 0 |
| | | Macro-average | - | - | .24 |
| GPT 3.5 Demo only for 2021 | .41 | Very worried | .40 | .49 | .34 |
| | | Somewhat worried | .53 | .39 | .86 |
| | | Not very worried | - | - | 0 |
| | | Not at all worried | - | - | 0 |
| | | Macro-average | - | - | .30 |

Supplemental Table 8. F1 Scores of GPTs for Presidential Election and Binary Belief in Global Warming across Subpopulations

| Variables | Election 2017 GPT4 Demo | | Election 2021 GPT4 Demo | | GW Belief Binary 2017 GPT4 Demo + Cov | | GW Belief Binary 2021 GPT4 Demo + Cov | |
|---|---|---|---|---|---|---|---|---|
| | F1 (Dem) | F1 (Rep) | F1 (Dem) | F1 (Rep) | F1 (Yes) | F1 (No) | F1 (Yes) | F1 (No) |
| Race/ethnicity | | | | | | | | |
| 2+ Races, Non-Hispanic<br>2017 EL *n* = 20<br>2021 EL *n* = 18<br>2017 GW *n* = 32<br>2021 GW *n* = 18 | 0.87 | 0.82 | 1 | 1 | 0.96 | 0.75 | 1 | 1 |
| Black, Non-Hispanic<br>2017 EL *n* = 89<br>2021 EL *n* = 67<br>2017 GW *n* = 104<br>2021 GW *n* = 73 | 0.96 | 0.25 | 0.96 | 0.55 | 0.95 | 0.29 | 0.96 | 0.25 |
| Hispanic<br>2017 EL *n* = 78<br>2021 EL *n* = 74<br>2017 GW *n* = 132<br>2021 GW *n* = 101 | 0.97 | 0.93 | 0.89 | 0.71 | 0.94 | 0.61 | 0.95 | 0.69 |
| Other, Non-Hispanic<br>2017 EL *n* = 30<br>2021 EL *n* = 31<br>2017 GW *n* = 51<br>2021 GW *n* = 38 | 1 | 1 | 0.96 | 0.88 | 0.98 | 0.83 | 0.94 | 0.33 |
| White, Non-Hispanic<br>2017 EL *n* = 668<br>2021 EL *n* = 614<br>2017 GW *n* = 786<br>2021 GW *n* = 669 | 0.9 | 0.92 | 0.9 | 0.89 | 0.92 | 0.72 | 0.94 | 0.78 |

Gender

    Female
    2017 EL *n* = 454
    2021 EL *n* = 403
    2017 GW *n* = 568
    2021 GW *n* = 454    0.94    0.91    0.91    0.87    0.93    0.66    0.95    0.75

    Male
    2017 EL *n* = 431
    2021 EL *n* = 401
    2017 GW *n* = 537
    2021 GW *n* = 445    0.9    0.91    0.91    0.89    0.93    0.74    0.94    0.76

Age

    18-29
    2017 EL *n* = 71
    2021 EL *n* = 84
    2017 GW *n* = 145
    2021 GW *n* = 113    0.93    0.89    0.96    0.93    0.95    0.75    0.94    0.5

    30-44
    2017 EL *n* = 172
    2021 EL *n* = 138
    2017 GW *n* = 243
    2021 GW *n* = 171    0.94    0.89    0.95    0.88    0.93    0.62    0.97    0.78

    45-59
    2017 EL *n* = 242
    2021 EL *n* = 214
    2017 GW *n* = 299
    2021 GW *n* = 230    0.92    0.9    0.85    0.85    0.93    0.63    0.93    0.75

    60+
    2017 EL *n* = 400
    2021 EL *n* = 368
    2017 GW *n* = 418
    2021 GW *n* = 385    0.92    0.93    0.91    0.89    0.93    0.76    0.94    0.79

Political ideology

    Moderate, middle
    of the road
    2017 EL *n* = 300
    2021 EL *n* = 303
    2017 GW *n* = 424
    2021 GW *n* = 380    0.88    0.82    0.86    0.74    0.94    0.54    0.96    0.68

| | | | | | | | | |
|---|---|---|---|---|---|---|---|---|
| Somewhat conservative<br>2017 EL *n* = 193<br>2021 EL *n* = 175<br>2017 GW *n* = 206<br>2021 GW *n* = 165 | 0.8 | 0.97 | 0.68 | 0.94 | 0.85 | 0.75 | 0.84 | 0.75 |
| Somewhat liberal<br>2017 EL *n* = 201<br>2021 EL *n* = 157<br>2017 GW *n* = 262<br>2021 GW *n* = 168 | 0.97 | 0.57 | 0.98 | 0.67 | 0.98 | 0.38 | 0.99 | 0.67 |
| Very conservative<br>2017 EL *n* = 99<br>2021 EL *n* = 87<br>2017 GW *n* = 99<br>2021 GW *n* = 84 | 0.73 | 0.98 | 0.55 | 0.97 | 0.61 | 0.82 | 0.77 | 0.87 |
| Very liberal<br>2017 EL *n* = 90<br>2021 EL *n* = 79<br>2017 GW *n* = 104<br>2021 GW *n* = 88 | 0.98 | 0.82 | 0.99 | - | 0.99 | 0.57 | 0.99 | - |
| Political party | | | | | | | | |
| Democrats<br>2017 EL *n* = 442<br>2021 EL *n* = 397<br>2017 GW *n* = 535<br>2021 GW *n* = 423 | 0.97 | 0.08 | 0.98 | - | 0.98 | 0.26 | 0.99 | 0.5 |
| Independent/Other<br>2017 EL *n* = 63<br>2021 EL *n* = 63<br>2017 GW *n* = 108<br>2021 GW *n* = 96 | 0.55 | 0.76 | 0.73 | 0.07 | 0.93 | 0.76 | 0.94 | 0.74 |
| No party/Not interested<br>2017 EL *n* = 27<br>2021 EL *n* = 8<br>2017 GW *n* = 105<br>2021 GW *n* = 50 | 0.64 | 0.69 | 0.55 | - | 0.88 | 0.58 | 0.89 | 0.67 |
| Republicans<br>2017 EL *n* = 353 | - | 0.97 | - | 0.94 | 0.85 | 0.77 | 0.86 | 0.79 |
| 2021 EL *n* = 334 | | | | | | | | |
| 2017 GW *n* = 354 | | | | | | | | |
| 2021 GW *n* = 321 | | | | | | | | |
| Education | | | | | | | | |
| Bachelor's degree or higher | | | | | | | | |
| 2017 EL *n* = 371 | | | | | | | | |
| 2021 EL *n* = 331 | | | | | | | | |
| 2017 GW *n* = 440 | | | | | | | | |
| 2021 GW *n* = 358 | 0.96 | 0.93 | 0.93 | 0.86 | 0.95 | 0.72 | 0.96 | 0.76 |
| High school | | | | | | | | |
| 2017 EL *n* = 213 | | | | | | | | |
| 2021 EL *n* = 195 | | | | | | | | |
| 2017 GW *n* = 272 | | | | | | | | |
| 2021 GW *n* = 226 | 0.88 | 0.89 | 0.86 | 0.86 | 0.91 | 0.7 | 0.93 | 0.72 |
| Less than high school | | | | | | | | |
| 2017 EL *n* = 34 | | | | | | | | |
| 2021 EL *n* = 25 | | | | | | | | |
| 2017 GW *n* = 64 | | | | | | | | |
| 2021 GW *n* = 42 | 0.9 | 0.83 | 0.9 | 0.86 | 0.93 | 0.56 | 0.99 | 0.94 |
| Some college | | | | | | | | |
| 2017 EL *n* = 267 | | | | | | | | |
| 2021 EL *n* = 253 | | | | | | | | |
| 2017 GW *n* = 329 | | | | | | | | |
| 2021 GW *n* = 273 | 0.91 | 0.92 | 0.92 | 0.91 | 0.92 | 0.71 | 0.93 | 0.76 |

Supplemental Table 9. F1 Scores of GPTs for Global Warming Belief, Cause, and Worry across Subpopulations

| Variables | GW Belief 2017 GPT4 Demo + Cov | | GW Belief 2021 GPT4 Demo + Cov | | GW Cause 2017 GPT4 Demo + Cov | | GW Cause 2021 GPT4 Demo + Cov | | GW Worry 2017 GPT4 Demo + Cov | | GW Worry 2021 GPT4 Demo + Cov | |
|---|---|---|---|---|---|---|---|---|---|---|---|---|
| | ACC | MA F1 | ACC | MA F1 | ACC | MA F1 | ACC | MA F1 | ACC | MA F1 | ACC | MA F1 |
| Race/ethnicity | | | | | | | | | | | | |
|   2+ Races, Non-Hispanic | .86 | - | .70 | - | .57 | - | .70 | - | .68 | .61 | .65 | - |
|   Black, Non-Hispanic | .83 | .48 | .81 | - | .53 | - | .59 | - | .53 | .48 | .57 | .57 |
|   Hispanic | .84 | - | .88 | - | .53 | - | .63 | - | .69 | .65 | .64 | .56 |
|   Other, Non-Hispanic | .79 | - | .95 | - | .62 | - | .90 | - | .60 | .54 | .64 | - |
|   White, Non-Hispanic | .75 | .54 | .81 | .58 | .52 | - | .60 | - | .67 | .67 | .55 | .53 |
| Gender | | | | | | | | | | | | |
|   Female | .78 | .53 | .85 | .60 | .52 | - | .61 | - | .65 | .65 | .58 | .54 |
|   Male | .77 | .54 | .79 | .53 | .54 | - | .62 | - | .66 | .65 | .56 | .54 |
| Age | | | | | | | | | | | | |
|   18-29 | .78 | .52 | .83 | .40 | .51 | - | .63 | - | .69 | .68 | .66 | .59 |
|   30-44 | .79 | .48 | .84 | .56 | .55 | - | .65 | - | .64 | .61 | .61 | .58 |
|   45-59 | .76 | .54 | .77 | .55 | .52 | - | .60 | - | .62 | .62 | .53 | .52 |
|   60+ | .77 | .56 | .84 | .61 | .54 | - | .60 | - | .67 | .68 | .55 | .52 |
| Political ideology | | | | | | | | | | | | |
| Moderate, middle of the road | .78 | .46 | .84 | .47 | .45 | - | .58 | - | .65 | .62 | .57 | .51 |
| Somewhat conservative | .62 | .50 | .65 | .50 | .43 | - | .42 | - | .61 | .60 | .52 | .51 |
| Somewhat liberal | .95 | - | .97 | .63 | .73 | - | .81 | - | .67 | .53 | .59 | - |
| Very conservative | .59 | .48 | .72 | .55 | .45 | - | .58 | - | .69 | .64 | .56 | .52 |
| Very liberal | .90 | - | .97 | - | .78 | - | .89 | - | .72 | .60 | .64 | - |
| Political party | | | | | | | | | | | | |
| Democrats | .94 | .44 | .95 | - | .69 | - | .80 | - | .70 | .57 | .61 | .44 |
| Independent/ Other | .71 | - | .77 | .46 | ..43 | - | .55 | - | .54 | .54 | .55 | .50 |
| No party/Not interested | .67 | .49 | .67 | .51 | .37 | - | .42 | - | .62 | .63 | .59 | .59 |
| Republicans | .63 | .51 | .72 | .57 | .42 | - | .45 | - | .65 | .62 | .52 | .50 |
| Education | | | | | | | | | | | | |
| Bachelor's degree or higher | .82 | .51 | .86 | .55 | .64 | - | .69 | - | .66 | .64 | .60 | .53 |
| High school | .71 | .53 | .79 | .54 | .45 | - | .53 | - | .64 | .65 | .53 | .51 |
| Less than high school | .76 | .54 | .74 | .58 | .43 | - | .54 | - | .64 | .61 | .63 | .61 |
| Some college | .77 | .56 | .80 | .59 | .48 | - | .59 | - | .66 | .65 | .56 | .55 |

Supplemental Table 10. Kullback-Leibler Distance

| Models | Presidential Election | Binary Global Warming Belief | Global Warming Belief | Causation of Global Warming | Global Warming worry |
|---|---|---|---|---|---|
| GPT 4.0 Demo + Cov for 2017 | - | .01 | .20 | .44 | .03 |
| GPT 4.0 Demo + Cov for 2021 | - | .02 | .09 | .40 | .01 |
| GPT 3.5 Demo + Cov for 2017 | - | .04 | .23 | 1.41 | .28 |
| GPT 3.5 Demo + Cov for 2021 | - | .06 | .19 | 1.19 | .22 |
| GPT 4.0 Demo only for 2017 | .0003 | .20 | 3.65 | 3.59 | 1.04 |
| GPT 4.0 Demo only for 2021 | .001 | .24 | 3.02 | 2.98 | 1.13 |
| GPT 3.5 Demo only for 2017 | .001 | .25 | 3.65 | 2.16 | 2.66 |
| GPT 3.5 Demo only for 2021 | .37 | .24 | 3.02 | 2.06 | 4.26 |

Note: Lower scores indicate a more similar distribution between a model and the survey. While row-by-row comparisons emphasize the similarity of a model's distribution to the survey within a specific column, direct column-by-column comparisons may be not impractical because of the different nature of the survey items.